# Agglomerative clustering and collectiveness measure via exponent generating function

REN Wei-Ya, LI Shuo-Hao, Guo Qiang; LI Guo-Hui; Zhang Jun

Email: weiyren.phd@gmail.com; lishuohao08@126.com; guoqiang05@nudt.edu.cn; gli2010a@163.com; zhangjun1975@nudt.edu.cn.

(College of Information System and Management, National University of Defense Technology, Hunan Changsha, 410073, China)

**Abstract:** The key in agglomerative clustering is to define the affinity measure between two sets. A novel agglomerative clustering method is proposed by utilizing a novel path integral descriptor. Firstly, the path integral descriptor of an edge, a node and a set are computed by path integral and exponent generating function. Then, the affinity measure between two sets is obtained by path integral descriptor of sets. Several good properties of the path integral descriptor is proposed in this paper. In addition, we give the physical interpretation of the proposed path integral descriptor of a set. The proposed path integral descriptor of a set can be regard as the collectiveness measure of a set, which can be a moving system such as human crowd, sheep herd and so on. Self-driven particle (SDP) model is used to test the ability of the proposed method in measuring collectiveness.

# 1 Introduction

Agglomerative clustering, which creates a hierarchy of clusters in a agglomerative process by merging small clusters or in a top-down. It begins with a large number of initial small clusters and iteratively select two clusters with the largest affinity under a certain affinity measure to merge [1][12][18][19].

The key in agglomerative clustering is the affinity measure between two sets. Since pairwise distances do not well capture the global structures of data, [1] propose a graph-structural agglomerative clustering, which utilized the path integral to define the affinity measure. The concept of path integral was first introduced in statistical mechanics and quantum mechanics [2][3][4], where it summed up the contributions of all possible paths to the evolution of a dynamical system. Path integral is one of the linkage algorithms in agglomerative clustering, which are free from the restriction on data distributions [20]. On the basis of path integral, [1] define a way to compute the path integral descriptor, which measures the stability of a set and is used to define the affinity measure between two sets.

[1] use a generating function to produce the path integral descriptor. Inspired by [1], we proposed a new method to compute the path integral descriptor of an edge, a node and a set, respectively. Exponent generating function is used to sum the path integral and get meaningful values. We show several good properties of the proposed path integral descriptor.

Large value of the path integral descriptor means that nodes in the set have a high coherence motion. Thus, the path integral descriptor is actually the measurement of collectiveness of a set. We use the integral descriptor as the collectiveness of a moving system, as same in [5]. Collectiveness of a set or a system is a meaningful work in social science and it is also useful in social psychology [15], control of swarming desert locusts[16], prevention of disease spreading [17], and many other fields. Collective motions such as fish shoal, sheep herd, and human crowd tend to follow the behaviors of others and move along the same direction as their neighbors [13][14].

[5] proposed a method which can quantify the structural properties of collective manifolds of crowds. The proposed collectiveness descriptor can also be used to quantify the structural properties of collective

manifolds of crowds. In addition, the proposed method has less parameters than method in [5]. We compare these two methods by self-driven particle (SDP) model [9].

The paper is organized as follows. The proposed agglomerative algorithm is proposed in Section 2. Then, the collectiveness measure is presented in Section 3. Conclusion is given in Section 4.

# 2 Agglomerative clustering

## 2.1 Neighborhood graph

Given a set $\mathcal{C}$ of samples $X = [x_1, x_2, ..., x_n] \in R^{D \times N}$, which $D$ is the dimension of data and $N = |\mathcal{C}|$ is the number of samples. Then, we build a directed graph $G = (V, E)$, where $V$ is the set of vertices corresponding to the samples in $X$, and $E$ is the set of edges connecting vertices. The graph is associated with a weighted adjacency matrix $W$, where $w_{ij} \in [0,1]$ is the weight of the edge from vertex $i$ to vertex $j$. $w_{ij} = 0$ if and only if there is no edge from $i$ to $j$. The famous $K$-NN graph is adopted as the edge connection strategy, which means each vertex has $K$ edges pointing from itself to its $K$ nearest neighbors.

## 2.2 Path integral and collectiveness via exponent generating function

The affinity measure between two clusters is the key of an agglomerative clustering algorithm [1]. Following [1], we compute affinity measure by the path integral descriptor.

### 2.2.1 Path integral

For simplicity, we consider all samples as a cluster set $\mathcal{C}$, and assume its weighted adjacency matrix is $W$. The path integral is a generalization of path counting, where paths in a cluster represent the connectivity [7]. We adopt the path integral definition in [1]: Computer path integral by selecting the starting and ending vertices of the path, let $\gamma_l(i_0, i_l) = \{i_0 \to i_1 \to i_2 \to \cdots \to i_l\}$ denote a directed path of length $l$ through nodes $i_0, i_1, ... i_l$ on $W$ between nodes $i_0$ and $i_l$. The path integral on a specific path $\gamma_l(i_0, i_l)$ [1] is defined as

$$\tau_{\gamma_l}(i_0, i_l) = \prod_{k=0}^{l} w(i_k, i_{k+1}). \tag{1}$$

Let the set $\mathcal{P}_l(i_0, i_l)$ denotes all the possible paths of length $l$ between nodes $i_0$ and $i_l$, then the $l$-path similarity $\tau_l(i_0, i_l)$ between nodes $i_0$ and $i_l$ is defined as [1]

$$\tau_l(i_0, i_l) = \sum_{\gamma_l(i_0, i_l) \in \mathcal{P}_l(i_0, i_l)} \tau_{\gamma_l}(i_0, i_l). \tag{2}$$

According to the algebraic graph theory [7], $\tau_l(i_0, i_l)$ can be efficiently computed as [1]

$$\tau_l(i_0, i_l) = W^l(i_0, i_l). \tag{3}$$

Then, $l$-path integral $\tau_l$ between all pairs of nodes can be computed as $\tau_l = W^l$ [7].

### 2.2.2 Generating function regularization

Give node $i$ and node $j$, we prefer to consider its all $l$-path integrals $\tau_l(i,j)$ ($l = 1,2,...,\infty$) by generating function regularization. Due to the scale of $W^l(i,j)$ become more and more large when $l$ become larger. We have to assign a meaningful value for the sum of a possibly divergent series [8] of $W^l(i,j)$ ($l = 1,2,...,\infty$) [1]. The value that consider all $l$-path integrals $\tau_l(i,j)$ ($l = 1,2,...,\infty$) can be regard as the integral descriptor of node $i$ and node $j$ [1]. Different from [1], we use the exponent generating function to define the path integral descriptor of an edge as follows

$$z(i,j) = \frac{1}{e^H}[\delta(i,j) + \sum_{l=1}^{\infty}\frac{W^l(i,j)}{l!}]. \quad (4)$$

where $\delta(i,j)$ is the Kronecker delta function defined as $\delta(i,j) = 1$ if $i = j$ and $\delta(i,j) = 0$ otherwise. $H = ||A||_\infty$ is a constant, where $A = [a]_{ij} = (W > 0)$, $A$ is a binary matrix that $a_{ij} = 1$ if $w_{ij} > 0$ and $a_{ij} = 0$ otherwise.

**Theorem 1** $z(i,j)$ is the $(i,j)$ entry of matrix $Z$, where $Z = \frac{1}{e^H}e^W$ and $0 \leq z(i,j) \leq \frac{1}{H} + \frac{\delta(i,j)}{e^H} - \frac{1}{He^H}$.

**Proof.**

From (4), we have

$$Z = \frac{1}{e^H}\left[I + \sum_{l=1}^{\infty}\frac{W^l}{l!}\right] = \frac{1}{e^H}e^W.$$

According to (4), it is obvious that $z(i,j) \geq 0$. Now we prove $z(i,j) \leq \frac{1}{H} + \frac{\delta(i,j)}{e^H} - \frac{1}{He^H}$. Given $A = (W > 0)$, then we have

$$z(i,j) = \frac{1}{e^H}\lim_{l\to\infty}\left[\delta(i,j) + \frac{W(i,j)}{1!} + \frac{W^2(i,j)}{2!} + \cdots + \frac{W^l(i,j)}{l!}\right]$$

$$\leq \frac{1}{e^H}\lim_{l\to\infty}\left[\delta(i,j) + \frac{A(i,j)}{1!} + \frac{A^2(i,j)}{2!} + \cdots + \frac{A^l(i,j)}{l!}\right] \triangleq \mathcal{M}(i,j).$$

Since $A = (W > 0)$ and $H = ||A||_\infty$, it means that the nonnegative values in each row of $A$ will not exceed $H$, we have $A^2(i,j) \leq H, A^3(i,j) \leq H^2,\ldots,A^l(i,j) \leq H^{l-1}$. We have

$$z(i,j) \leq \mathcal{M}(i,j) \leq \frac{1}{e^H}\lim_{l\to\infty}\left[\delta(i,j) + \frac{1}{1!} + \frac{H}{2!} + \cdots + \frac{H^{l-1}}{l!}\right]$$

$$= \frac{1}{He^H}\lim_{l\to\infty}\left[\delta(i,j)H - 1 + 1 + \frac{H}{1!} + \frac{H^2}{2!} + \cdots + \frac{H^l}{l!}\right] = \frac{1}{He^H}[\delta(i,j)H - 1 + e^H]$$

$$= \frac{1}{H} + \frac{\delta(i,j)}{e^H} - \frac{1}{He^H}. \qquad \blacksquare$$

### 2.2.2 Path integral descriptor

After defining the path integral descriptor of an edge, we can define the path integral descriptor of a node and a set by using method in [1]. Path integral descriptor of node $i$ at $l$-path scale is defined as

$$\varphi_l(i) = \sum_{j\in\mathcal{C}}\tau_l(i,j) = [W^l\mathbf{1}]_i. \quad (5)$$

where $\mathbf{1} \in R^{|\mathcal{C}|\times 1}$ is a vector with all elements as one, $[\cdot]_i$ denotes $i$-th element of a vector.

Then the path integral descriptor of set at $l$-path scale is defined as the mean of all node collectiveness [1]

$$\Phi_l = \frac{1}{|\mathcal{C}|}\sum_{i=1}^{|\mathcal{C}|}\varphi_l(i) = \frac{1}{|\mathcal{C}|}\mathbf{1}^T W^l \mathbf{1}. \quad (6)$$

Path integral descriptor of node on all the path integral is written as [1]

$$\varphi(i) = \frac{1}{e^H}[1 + \sum_{l=1}^{\infty}\frac{\varphi_l(i)}{l!}] = [Z\mathbf{1}]_i. \quad (7)$$

Path integral descriptor of set [1] on all the path integral is written as

$$\Phi = \frac{1}{|\mathcal{C}|}\sum_{i=1}^{|\mathcal{C}|}\varphi(i) = \frac{1}{|\mathcal{C}|}\mathbf{1}^T Z \mathbf{1} = \frac{1}{|\mathcal{C}|e^H}\mathbf{1}^T e^W \mathbf{1}. \qquad (8)$$

Then, one can denote the path integral descriptor of any set by (8). In this instance, $\Phi = \Phi_{\mathcal{C}}$ for set $\mathcal{C}$. In section 3, we will further explain the path integral descriptor in physical intuition.

**Property 1 (Convergence).** $Z$ *always converges.*
**Proof.** Since $0 \leq W_{ij} \leq 1$, we have $W \leq U$ by defining a matrix $U \in R^{|\mathcal{C}|\times|\mathcal{C}|}$ with all elements as one. $Z = \frac{1}{e^H}e^W \leq \frac{1}{e^H}e^U$. Thus, $Z$ always converges. ∎

**Property 2.** *If $W_{i,i} = 0$, then $\Phi(W) = \Phi(W + I)$. If $W + I$ is also block diagonal then $\Phi(W) \leq 1$. The equality only stands when $W = (W > 0)$ and all blocks have same size.*
**Proof**

Given $A = (W > 0)$ and $H = ||A||_\infty$. Since $W_{i,i} = 0$, then $||A + I||_\infty = H + 1$, we have $Z(W) = \frac{e^W}{e^H}$ and $Z(W + I) = \frac{e^{W+I}}{e^{H+1}}$. It is easy to know that $\frac{Z(W)}{Z(W+I)} = 1$, thus $Z(W) = Z(W + I)$ and

$$\Phi(W) = \frac{1}{|\mathcal{C}|}\mathbf{1}^T Z(W)\mathbf{1} = \frac{1}{|\mathcal{C}|}\mathbf{1}^T Z(W+I)\mathbf{1} = \Phi(W + I).$$

If $W + I$ is block diagonal, and has $c$ blocks with size of $k_1, k_1, \ldots, k_c$, respectively. We have

$$(A+I)^l(i,j) = \begin{cases} k_i^{l-1}, & \text{if } (i,j) \text{ belongs to block } i,\ i \in \{1,2,\ldots,c\} \\ 0, & \text{otherwise.} \end{cases}$$

Since $H + 1 = ||A + I||_\infty \geq k_i \geq 1, i \in \{1,2,\ldots,c\}$, $\sum_i k_i = |\mathcal{C}|$, we have

$$\Phi(W) = \Phi(W + I) = \frac{1}{|\mathcal{C}|}\mathbf{1}^T Z(W+I)\mathbf{1} = \frac{1}{|\mathcal{C}|}\mathbf{1}^T \frac{e^{W+I}}{e^{H+1}}\mathbf{1}$$

$$= \frac{1}{|\mathcal{C}|e^{H+1}}\mathbf{1}^T\left[I + (W+I) + \frac{(W+I)^2}{2!} + \cdots + \frac{(W+I)^\infty}{\infty!}\right]\mathbf{1}$$

$$\leq \frac{1}{|\mathcal{C}|e^{H+1}}\mathbf{1}^T\left[I + (A+I) + \frac{(A+I)^2}{2!} + \cdots + \frac{(A+I)^\infty}{\infty!}\right]\mathbf{1}$$

$$= \frac{1}{|\mathcal{C}|e^{H+1}}\left[|\mathcal{C}| + \sum_i k_i^0 * k_i^2 + \frac{\sum_i k_i^1 * k_i^2}{2!} + \cdots + \frac{\sum_i k_i^\infty * k_i^2}{\infty!}\right]$$

$$\leq \frac{1}{|\mathcal{C}|e^{H+1}}\left[|\mathcal{C}| + (H+1)\sum_i k_i + \frac{(H+1)^2 \sum_i k_i}{2!} + \cdots + \frac{(H+1)^\infty \sum_i k_i}{\infty!}\right]$$

$$= \frac{|\mathcal{C}|}{|\mathcal{C}|e^{H+1}}e^{H+1} = 1.$$

The equality only stands when $W = (W > 0)$ and all blocks have same size such that $k_1 = k_2 = \cdots = k_c = H + 1$. ∎

**Property 3 (Bounds of $\Phi$).** $0 \leq \Phi \leq 1$.
**Proof.**

Since we have $w_{ij} \in [0,1]$ and $W\mathbf{1} \leq A\mathbf{1} \leq H\mathbf{1}$ by giving $A = (W > 0)$ and $H = ||A||_\infty$. Then,

$$\Phi = \frac{1}{|\mathcal{C}|e^H} \mathbf{1}^T e^W \mathbf{1} = \frac{1}{|\mathcal{C}|e^H} \mathbf{1}^T \left[ I + \sum_{l=1}^{\infty} \frac{W^l}{l!} \right] \mathbf{1} = \frac{1}{|\mathcal{C}|e^H} \lim_{l \to \infty} [\mathbf{1}^T \mathbf{1} + \mathbf{1}^T W \mathbf{1} + \frac{\mathbf{1}^T W^2 \mathbf{1}}{2!} + \cdots + \frac{\mathbf{1}^T W^l \mathbf{1}}{l!}]$$

$$\leq \frac{1}{|\mathcal{C}|e^H} \lim_{l \to \infty} \left[ \mathbf{1}^T \mathbf{1} + H \mathbf{1}^T \mathbf{1} + \frac{H^2 \mathbf{1}^T \mathbf{1}}{2!} + \cdots + \frac{H^l \mathbf{1}^T \mathbf{1}}{l!} \right]$$

$$\leq \frac{|\mathcal{C}|}{|\mathcal{C}|e^H} \lim_{l \to \infty} \left[ 1 + H + \frac{H^2}{2!} + \cdots + \frac{H^l}{l!} \right] = \frac{|\mathcal{C}|}{|\mathcal{C}|e^H} e^H = 1. \qquad \blacksquare$$

**Property 4 (Bounds of $\Phi_l$).** $0 \leq \Phi_l \leq H^l$.

**Proof.** (same method in [5], but we use the property $H = ||A||_\infty$)

It is obvious that $\Phi_l = \frac{1}{|\mathcal{C}|} \mathbf{1}^T W^l \mathbf{1} \geq 0$. Now we prove $\Phi_l \leq H^l$. Given $A = (W > 0)$ and $H = ||A||_\infty$, we have $W\mathbf{1} \leq A\mathbf{1} \leq H\mathbf{1}$. We use mathematical induction to prove the statement $\mathbf{1}^T W^n \mathbf{1} \leq |\mathcal{C}|H^n$ holds for any $n$ as follows:

(1) When $n=1$, since $w(i,j) \leq a(i,j) = 1$, we have

$$\mathbf{1}^T W \mathbf{1} \leq \sum_{i,j} w(i,j) \leq \sum_{i,j} a(i,j) \leq \mathbf{1}^T A \mathbf{1} \leq |\mathcal{C}|H.$$

(2) When $n = m$, we assume the statement $\mathbf{1}^T W^m \mathbf{1} \leq |\mathcal{C}|H^m$, and let $w_m(i,j)$ denote the $(i,j)$ entry of $W^m$, then

$$\mathbf{1}^T W^{m+1} \mathbf{1} = \mathbf{1}^T W^m W \mathbf{1} = \sum_{i,j,k} w_m(i,k) w(k,j) \leq \sum_{i,j,k} w_m(i,k) a(k,j) = \mathbf{1}^T W^m A \mathbf{1} \leq \mathbf{1}^T W^m H \mathbf{1}$$

$$\leq |\mathcal{C}|H^{n+1}.$$

Then the statement $\mathbf{1}^T W^{m+1} \mathbf{1} \leq |\mathcal{C}|H^{m+1}$ holds. Thus $\mathbf{1}^T W^n \mathbf{1} \leq |\mathcal{C}|H^n$ holds for any $n$. $\blacksquare$

**Property 5 (Asymptotic limit of $\Phi_l$).** $\lim_{l \to \infty} \frac{\ln \Phi_l}{l} = \ln \lambda \leq \ln H$, where $A = (W > 0)$, $H = ||A||_\infty$, and $\lambda$ is the largest eigenvalue of $W$.

**Proof.** (same method in [5], but we use the property $H = ||A||_\infty$)

Since $w_{ii} = 0$, then $a_{ii} = 0$. Denote $\lambda_{i=1,2,\ldots,|\mathcal{C}|}$ are eigenvalues of $W$. Assume $\lambda$ is the $i$-th eigenvalue $\lambda_i$. According to Gershgorin Circle Theorem, we have

$$\lambda \leq |\lambda_i| \leq \sum_{j \neq i} |a_{ij}| \leq ||A||_\infty = H.$$

Then $\ln \lambda \leq \ln H$.

According to the Perron-Frobenius theorem, we have $\lim_{l \to \infty} \frac{W^l}{\lambda^l} = vw^T$, where $v$ and $w$ are left and right eigenvectors of $W$ corresponding to $\lambda$ and are normalized to $w^T v = 1$. Then we have

$$\lim_{l \to \infty} \frac{\ln \Phi_l}{l} - \ln \lambda = \lim_{l \to \infty} \frac{\ln \frac{\Phi_l}{\lambda^l}}{l} = \lim_{l \to \infty} \frac{\ln \frac{\mathbf{1}^T W^l \mathbf{1}}{|\mathcal{C}|\lambda^l}}{l} = \lim_{l \to \infty} \frac{\ln \frac{\mathbf{1}^T vw^T \mathbf{1}}{|\mathcal{C}|}}{l} = 0.$$

Thus $\lim_{l \to \infty} \frac{\ln \Phi_l}{l} = \ln \lambda \leq \ln H$. $\blacksquare$

It also shows the exponential growth of $\Phi_l$ with $l$.

**Property 6 (Approximate error bound of $Z$).** $||Z - Z_{1 \sim n}|| \leq \frac{1}{e^H} \sum_{l=n+1}^{D-1} \frac{W^l}{l!} + \frac{1}{e^H} \frac{||W||^D}{D!} \frac{D+1}{D+1-||W||}$ if

$n < D - 2$, and $||Z - Z_{1 \sim n}|| \leq \frac{1}{e^H} \frac{||W||^{(n+1)}}{(n+1)!} \frac{n+2}{n+2-||W||}$ otherwise. $Z_{1 \sim n}$ ($n$) denotes the sum of first $n$ terms of $Z$, the matrix norm of matrix $B$ ($b_{ij} \in [0,1]$) is defined as $||B|| = \sum_{ij} B_{ij}$, and $||A|| = D > 0$.

**Proof.** (See approximate error bound details in [21])

If $n < D - 2$, let $R = Z - Z_{1 \sim n} - \frac{1}{e^H} \sum_{l=n+1}^{D-1} \frac{W^l}{l!} = \frac{1}{e^H} \sum_{l=D}^{\infty} \frac{W^l}{l!}$. We can find that $||B_1 B_2|| \leq ||B_1|| \cdot ||B_2||$, where $B_1, B_2$ are two matrices that range from $[0,1]$. Thus, $||W^l|| \leq ||W||^l$. Notice that $||W|| \leq ||A|| = D \leq |\mathcal{C}|H$, then we have

$$||R|| = \frac{1}{e^H} \sum_{l=D}^{\infty} \frac{||W^l||}{l!} \leq \frac{1}{e^H} \sum_{l=D}^{\infty} \frac{||W||^l}{l!} = \frac{1}{e^H} \lim_{l \to \infty} \left[ \frac{||W||^D}{D!} + \frac{||W||^{(D+1)}}{(D+1)!} + \cdots + \frac{||W||^l}{l!} \right] =$$

$$= \frac{1}{e^H} \frac{||W||^D}{D!} \lim_{l \to \infty} \left[ 1 + \frac{||W||}{D+1} + \cdots + \frac{||W||^{(l-D)}}{(D+1) \ldots (l-1)l} \right]$$

$$\leq \frac{1}{e^H} \frac{||W||^D}{D!} \lim_{l \to \infty} \left[ 1 + \frac{||W||}{D+1} + \cdots + \frac{||W||^{(l-D)}}{(D+1)^{(l-D)}} \right] = \frac{1}{e^H} \frac{||W||^D}{D!} \frac{1}{1 - ||W||/(D+1)}$$

$$= \frac{1}{e^H} \frac{||W||^D}{D!} \frac{D+1}{D+1-||W||}.$$

If $n \geq D - 2$, let $R = Z - Z_{1 \sim n} = \frac{1}{e^H} \sum_{l=n+1}^{\infty} \frac{W^l}{l!}$. Then we can find that

$$||R|| \leq = \frac{1}{e^H} \frac{||W||^{(n+1)}}{(n+1)!} \frac{n+2}{n+2-||W||}. \qquad \blacksquare$$

## 2.3 Why $e^W$ ?

**Property 7 (Bounds of $W^l$).** $0 \leq W^l(i,j) \leq H^{l-1}$, $l = 1,2,\ldots,\infty$.

**Proof** (same method in [5], but we use the property $H = ||A||_\infty$)

Obviously, $W^l(i,j) \geq 0$ since $0 \leq W(i,j) \leq 1$. Given $A = (W > 0)$, it is easy to know that $W^l(i,j) \leq A^l(i,j)$, $l = 1,2,\ldots,\infty$. Denote $H = ||A||_\infty$, thus the nonnegative values in each row of $A$ will not exceed $H$. Then, we have $A^2(i,j) \leq H$, $A^3(i,j) \leq H^2, \ldots, A^l(i,j) \leq H^{l-1}$. Finally,
$$W^l(i,j) \leq A^l(i,j) \leq H^{l-1}, \qquad l = 1,2,\ldots,\infty. \qquad \blacksquare$$

We rewrite $Z$ as follows

$$Z = \frac{1}{e^H} e^W = \frac{1}{e^H} \left[ 1 + W + \frac{W^2}{2!} + \cdots + \frac{W^\infty}{\infty!} \right] = \alpha_0 I + \alpha_1 W + \alpha_2 W^2 + \cdots + \alpha_\infty W^\infty. \qquad (9)$$

Thus, $Z$ can be seen as the linear combination of $I, W, W^2, \ldots, W^\infty$. Obviously, $\alpha_l = \frac{1}{l! e^H}$ and $\alpha_0 = \alpha_1 > \alpha_2 > \cdots > \alpha_\infty$. It seems $Z$ pay less and less attention to $W^l$ when $l$ become larger. However, $W, W^2, \ldots, W^\infty$ take values from different scales. If we normalized $W^l$ by making $W^l(i,j)$ ranges from $[0,1]$, i.e., $\widetilde{W}^l = W^l / H^{l-1}$. We have

$$Z = \alpha_0 I + \alpha_1 W + \alpha_2 W^2 + \cdots + \alpha_\infty W^\infty = \tilde{\alpha}_0 I + \tilde{\alpha}_1 \widetilde{W} + \tilde{\alpha}_2 \widetilde{W}^2 + \cdots + \tilde{\alpha}_\infty \widetilde{W}^\infty. \qquad (10)$$

where $\tilde{\alpha}_0 = \frac{1}{e^H}$, and $\tilde{\alpha}_l = \frac{H^{l-1}}{l! e^H}$, $l = 1,2,\ldots,\infty$.

**Theorem 2** $\frac{1}{e^H} \leq \tilde{\alpha}_l \leq \frac{H^{H-1}}{H! e^H}$ for $l = 1,2,\ldots,H$. The left equality stands when $l = 1$ while the right equality stands when $l = H$ or $l = H - 1$. $0 \leq \tilde{\alpha}_l < \frac{H^{H-1}}{(H)! e^H}$ for $l = H+1, H+2, \ldots, \infty$.

**Proof**

When $l \leq H$, $\frac{\tilde{\alpha}_l}{\tilde{\alpha}_{l-1}} = \frac{H^{l-1}}{l!e^H}\frac{(l-1)!e^H}{H^{l-2}} = \frac{H}{l} \geq 1$. The equality only stands when $l = H$. Thus, $\tilde{\alpha}_H \geq \tilde{\alpha}_{H-1} > \cdots > \tilde{\alpha}_1 = \frac{1}{e^H}$.

When $l \geq H - 1$, $\frac{\tilde{\alpha}_l}{\tilde{\alpha}_{l+1}} = \frac{H^{l-1}}{l!e^H}\frac{(l+1)!e^H}{H^l} = \frac{l+1}{H} \geq 1$. The equality only stands when $l = H - 1$. Thus, $\tilde{\alpha}_{H-1} \geq \tilde{\alpha}_H > \cdots > \tilde{\alpha}_\infty = 0$.

Then, we have $\tilde{\alpha}_H = \tilde{\alpha}_{H-1} > \tilde{\alpha}_l$, $l = 1,2,\ldots,\infty$ and $l \neq H, H-1$. ∎

From Theorem 2, we know that $Z$ be seen as the linear combination of $I, \widetilde{W}, \widetilde{W}^2, \ldots, \widetilde{W}^\infty$. $\tilde{\alpha}_H$ and $\tilde{\alpha}_{H-1}$ are two largest values among all coefficients. If $\widetilde{W}$ is a $K$-NN graph, then $H = ||A||_\infty = K$ and each node points to $K$ neghbors. It means that, $Z$ pay more attention to $\widetilde{W}^H, \widetilde{W}^{H-1}$ and other $\widetilde{W}^l$ if $l$ is close to $H-1$ or $H$, i.e., $Z$ let a node pay more attention to nodes which are around the boundary of it. Thus, nodes that close to the boundary of a node determine which cluster it belongs to. We show an example in Fig. 3. The value of $\widetilde{W}^l$ will also affect the value of $\tilde{\alpha}_l \widetilde{W}^l$. If $\widetilde{W}^l = 0$, we have $\tilde{\alpha}_l \widetilde{W}^l = 0$ even when $l$ is close to $H - 1$ or $H$. A certain $l$ will make $||\tilde{\alpha}_l \widetilde{W}^l||_2$ take the largest value and $l < H$. The value of $\widetilde{W}^l$ will be small only when the set has little strong connected edges, which means relative far nodes from each node are unreliable. It also means that if the $K$-NN graph is unreliable, node will consider nodes that are relative close nodes as the boundary of it.

## 2.4 Agglomerative clustering

We adopt the agglomerative clustering procedure in [1], initial small clusters are simply constructed as *weakly connected components* [6] of a $K_o$-NN graph ($K_o$ is small, typically as 1 or 2) [1].

Then, the structural affinity between pair of sets should be defined on the $K$-NN graph ($K$ is larger than $K_o$) to decide the merging strategy. Only two clusters at every iteration is merged when they have the maximum structural affinity [1].

Set $\Gamma_a$'s path integral descriptor can be written as $\Phi_{\Gamma_a}$. According to [1], if the nodes in two sets $\Gamma_a, \Gamma_b$ are strongly connected, merging them will create many new edges in $\Gamma_a$, and also for $\Gamma_a$. More edges in $\Gamma_a$ means the value of $\Phi_{\Gamma_a}$ will be larger. Thus, the structural affinity can be defined by path integral descriptor of two sets [1]. We adopt the structural affinity definition in [1], and define the structural affinity $\mathcal{A}_{\Gamma_a,\Gamma_b}$ of set $\Gamma_a$ and set $\Gamma_b$ as

$$\mathcal{A}_{\Gamma_a,\Gamma_b} = (\Phi_{\Gamma_a|\Gamma_a \cup \Gamma_b} - \Phi_{\Gamma_a}) + (\Phi_{\Gamma_b|\Gamma_a \cup \Gamma_b} - \Phi_{\Gamma_b}). \tag{11}$$

where $\Phi_{\Gamma_a|\Gamma_a \cup \Gamma_b}$ is the conditional path integral descriptor, which means all the paths to compute $\Phi$ could lie in $\Gamma_a \cup \Gamma_b$ but their starting and ending nodes must be within $\Gamma_a$.

From formula (8), the key contribution of this paper is to define $\Phi$ as follows

$$\Phi_{\Gamma_a} = \frac{1}{|\Gamma_a|e^{H_{\Gamma_a}}} \mathbf{1}_{\Gamma_a}^T \mathbf{1}^T e^{W_{\Gamma_a}} \mathbf{1}_{\Gamma_a}, \quad \Phi_{\Gamma_b} = \frac{1}{|\Gamma_b|e^{H_{\Gamma_b}}} \mathbf{1}_{\Gamma_b}^T \mathbf{1}^T e^{W_{\Gamma_b}} \mathbf{1}_{\Gamma_b}. \tag{12}$$

where $|\Gamma_a|, |\Gamma_b|$ are the number of nodes in $\Gamma_a, \Gamma_b$, respectively. $\mathbf{1}_{\Gamma_a} \in R^{|\Gamma_a| \times 1}$, $\mathbf{1}_{\Gamma_b} \in R^{|\Gamma_b| \times 1}$ are vectors with all elements as one. $H_{\Gamma_a} = ||(W_{\Gamma_a} > 0)||_\infty$ and $H_{\Gamma_b} = ||(W_{\Gamma_b} > 0)||_\infty$, where $W_{\Gamma_a}$ and $W_{\Gamma_b}$ are the sub-graphs of $W_a$ respecting to nodes set $\Gamma_a, \Gamma_b$, respectively.

$$\Phi_{\Gamma_a|\Gamma_a \cup \Gamma_b} = \frac{1}{|\Gamma_a|e^{H_{\Gamma_a|\Gamma_a \cup \Gamma_b}}} \mathbf{1}_{\Gamma_a|\Gamma_a \cup \Gamma_b}^T e^{W_{\Gamma_a|\Gamma_a \cup \Gamma_b}} \mathbf{1}_{\Gamma_a|\Gamma_a \cup \Gamma_b}. \tag{13}$$

where $H_{\Gamma_a|\Gamma_a \cup \Gamma_b} = ||(W_{\Gamma_a|\Gamma_a \cup \Gamma_b} > 0)||_\infty$ and $W_{\Gamma_a|\Gamma_a \cup \Gamma_b}$ means a graph that edges may lie in $\Gamma_a \cup \Gamma_b$ but nodes only lie in $\Gamma_a$. $\mathbf{1}_{\Gamma_a|\Gamma_a \cup \Gamma_b} \in R^{(|\Gamma_a|+|\Gamma_b|) \times 1}$ is the vector in which the elements corresponding to the vertices in $\Gamma_a$ are all one and other elements are zero.

$$\Phi_{\Gamma_b|\Gamma_a\cup\Gamma_b} = \frac{1}{|\Gamma_b|e^{H_{\Gamma_b|\Gamma_a\cup\Gamma_b}}} \mathbf{1}^T_{\Gamma_b|\Gamma_a\cup\Gamma_b} e^{W_{\Gamma_b|\Gamma_a\cup\Gamma_b}} \mathbf{1}_{\Gamma_b|\Gamma_a\cup\Gamma_b}. \tag{14}$$

where $H_{\Gamma_b|\Gamma_a\cup\Gamma_b} = ||(W_{\Gamma_b|\Gamma_a\cup\Gamma_b} > 0)||_\infty$ and $W_{\Gamma_b|\Gamma_a\cup\Gamma_b}$ means a graph that edges may lie in $\Gamma_a \cup \Gamma_b$ but nodes only lie in $\Gamma_b$. $\mathbf{1}_{\Gamma_b|\Gamma_a\cup\Gamma_b} \in R^{(|\Gamma_a|+|\Gamma_b|)\times 1}$ is the vector in which the elements corresponding to the vertices in $\Gamma_b$ are all one and other elements are zero.

According to [1], when the agglomerative clustering stops, assume we get $k$ clusters denoted as $\Gamma'_c (c = 1,2, ... , k)$. Then the exemplar of cluster $\Gamma'_1, \Gamma'_2, ... , \Gamma'_k$ can be found [1]. For cluster $\Gamma'_c$, [1] suggest one can find the exemplar of it by solving

$$exemplar_{\Gamma'_c} \triangleq \underset{i\in\Gamma'_c}{argmax} \sum_j \Phi_{\Gamma'_c}(i,j) + \sum_j \Phi_{\Gamma'_c}(j,i). \tag{15}$$

From the above formula, node $i \in \Gamma'_c$ will have the largest contribution in $\Phi_{\Gamma'_c}$ when node $i$ is the exemplar of cluster $\Gamma'_c$, which is reasonable [1].

The overall agglomerative algorithm is presented in Algorithm 1.

If we want to automatically determine the number of clusters, the path integral descriptor can considered. Assume data has $k$ clusters and there are no edges between different clusters, the weighted matrix $W$ will be block diagonal through permutation. Property 2 shows that if the weighted matrix is block diagonal, its path integral descriptor equals one when elements in $W$ are binary form and all clusters have the same size. We can estimate the number of clusters by considering the path integral descriptor. For example, we can build a new graph based on the set partition after each iteration in algorithm1 and compute the path integral descriptor of it. In addition, if $\mathcal{A}_{\Gamma_a,\Gamma_b}$ do not increase anymore, the number of clusters could be obtain. However, the accurate determination of cluster numbers is a difficult problem and also a specific research topic in the field of pattern clustering.

**Algorithm.1 Agglomerative Clustering via exponent generating function (algorithm adopt the procedure in [1])**

---
**Input**: Dataset $X$ and the target number of clusters $k$. Form weighted matrix $W$ by $K$-NN graph. Form $k_0$ initial clusters $\Gamma' = \{\Gamma_c, c = 1,2, ... , k_0\}$ by $K_0$–NN graph.
**while** $k_0 \geq k$ **do**
  Search two clusters $\Gamma_a, \Gamma_b$, such that $\{\Gamma_a, \Gamma_b\} = argmax_{\{\Gamma_a,\Gamma_b\}\subset\Gamma} \mathcal{A}_{\Gamma_a,\Gamma_b}$, where $\mathcal{A}_{\Gamma_a,\Gamma_b}$ is defined in (11).
  $\Gamma' \leftarrow \{\Gamma' \setminus \{\Gamma_a \cap \Gamma_b\}\} \cup \{\Gamma_a \cap \Gamma_b\}$, and $k_0 = k_0 - 1$.
**end while**
  Find the exemplar point in each cluster of $\Gamma'$ by (15), denoted as $exemplar_{\Gamma'_c}, c = 1,2, ... , k$.
**output**: $\Gamma$, $exemplar_{\Gamma'_c}, c = 1,2, ... , k$.

---

# 3 Collectiveness

In section 2.2, we introduce the path integral descriptor $\Phi$ of a set. If nodes in a set have a high coherence, $\Phi$ will have a large value. If nodes in a set are similar with each other, $\Phi$ will also be large. Thus, the path integral descriptor can be comprehended as the measurement of collectiveness of the set. In fact, reference [5] already uses this idea. However, the difference between the proposed paper and [5] is the different method to compute $\Phi$.

The self-driven particle (SDP) model [9] is a famous model for studying collective motion and shows high similarity with various crowd systems in nature [10][11]. In this section, we use this model to test the proposed path integral descriptor, or named collectiveness, same as [5]. The ground-truth of collectiveness in SDP is known for evaluation. SDP model produce a system of moving particles that are driven with a constant speed [5]. SDP gradually turns into collective motion from disordered motion [5]. Each particle will update its direction of velocity to the average direction of the particles in its neighborhood at each frame [5].

The update of velocity direction $\theta$ [9] for every particles $i$ in SDP is

$$\theta_i(t+1) = <\theta_j(t)>_{j \in \mathcal{N}(i)} + \Delta\theta. \quad (16)$$

where $<\theta_j(t)>_{j \in \mathcal{N}(i)}$ denotes the average direction of velocities of particles within the neighborhood $\mathcal{N}(i)$ of $i$. $\Delta\theta$ is a random angle chosen with a uniform distribution within the interval $[-\eta\pi, \eta\pi]$, where $\eta$ tunes the noise level of alignment [9].

## 3.1 Neighborhood graph

Given $N$ moving particles in SDP, we can measure the similarity of particles by $K$-NN graph. For simplicity, we adopt the method in [5] to compute the $K$-NN graph for comparison. At time $t$, the weight value on edge between particle $i$ and particle $j$ are defined by [5]

$$w_t(i,j) = \begin{cases} max(\frac{v_i v_j^T}{||v_i||_2 ||v_j||_2}, 0), & if\ j \in \mathcal{N}(i) \\ 0, & otherwise \end{cases}. \quad (17)$$

where $v_i$ is the velocity vector of particle $i$.

## 3.2 Numerical analysis

We compare the proposed method with the state-of-the-art method in [5]. As seen in Fig.1, we show an example of measuring collective motion in SDP.

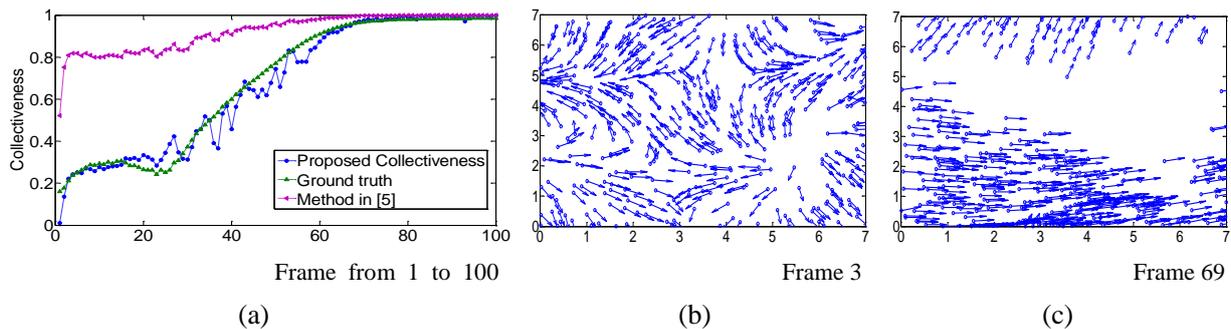

(a)      (b)      (c)

Fig. 1 An example of measuring collective motion in SDP, and $N = 400$, $K = 20$, size of ground $L = 7$, the absolute value of velocity $||v|| = 0.03$, the interaction radius $r = 1$ and $\eta = 0$. At the beginning, collectiveness $\Phi$ is low since the spatial locations and moving directions of individuals are randomly assigned. The behaviors of individuals gradually turn into collective motion from random movements. (a) Compare the collectiveness ground-truth with collectiveness measured by the proposed method and by [5]. (b) Frame 3, the collectiveness of ground-truth, method in [5] and the proposed method are 0.05, 0.77 and 0.16, respectively. (c) Frame 69, the collectiveness of ground-truth, method in [5] and the proposed method are 0.91, 0.98 and 0.93, respectively.

We compute the collectiveness of all frames and then compute the relevant coefficient between the measured collectiveness and the ground truth. As seen in Fig. 2, we show an example that analyze the relevant between the ground truth (GT) and the method in [5] and the proposed method, and the proposed method has a higher relevant coefficient value. We also show the average relevant coefficient between the ground truth (GT) and two methods (100 runs).

In Fig. 3, we analyze the components of $Z = \frac{e^W}{e^H} = \frac{1}{e^H}\sum_{l=1}^{\infty}\frac{W^l}{l!}$. We draw normalized value of $||\frac{1}{e^H}\frac{W^l}{l!}||_2$ with respect to $l = 1, \ldots, 100$.

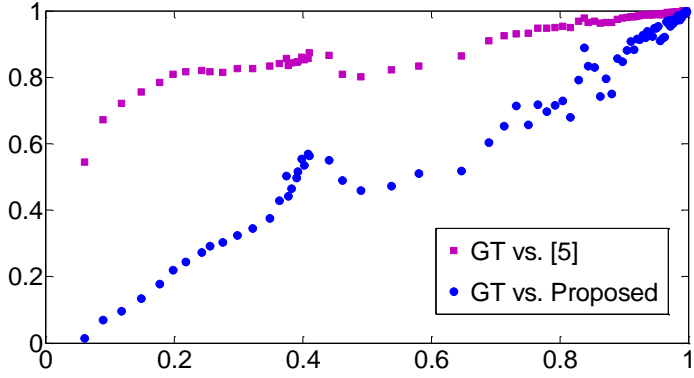

Fig. 2 An example. Relevant analysis of the measured collectiveness and the ground truth collectives (system order). The relevant coefficient between the ground truth (GT) and the method in [5] is 0.88, while the relevant coefficient between the ground truth (GT) and the proposed method 0.94. Parameters: $N = 400$, $K = 20$, size of ground $L = 7$, the absolute value of velocity $||v|| = 0.03$, the interaction radius $r = 1$ and $\eta = 0$

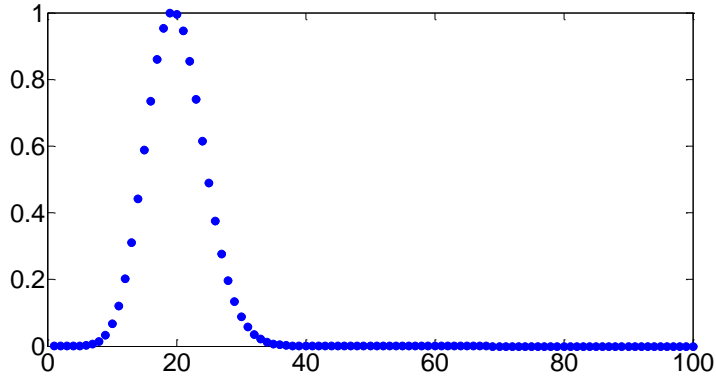

Fig. 3 Draw normalized value of $||\frac{1}{e^H}\frac{W^l}{l!}||_2, l = 1,…,100$. $||\frac{1}{e^H}\frac{W^l}{l!}||_2$ has largest value when $l = 19$ ($K = 20$). Parameters: $N = 400$, $K = 20$, size of ground $L = 7$, the absolute value of velocity $||v|| = 0.03$, the interaction radius $r = 1$ and $\eta = 0$.

Tab.1 Average relevant coefficient between the ground truth (GT) and two methods (100 runs, 100 frames in each run). Fix $K = 20$, and $||v|| = 0.03$, $r = 1$, $L = 7$ and $\eta = 0$.

| Relevant Coefficient | GT vs. [5] | GT vs. proposed |
|---|---|---|
| $N = 200$ | 0.86 | **0.90** |
| $N = 400$ | 0.81 | **0.88** |
| $N = 500$ | 0.83 | **0.92** |

## 4 Conclusion

In this paper, we define the path integral descriptor of an edge, a node and a set by the exponent generating function on the basis of the path integral and [1], respectively. Several fine properties of the path integral descriptor have been shown. The proposed path integral descriptor is used to define the structure affinity to determine how to emerge sets in agglomerative clustering, and then we follow the agglomerative clustering procedure in [1] to propose a agglomerative clustering algorithm. We regard the path integral descriptor as the collectiveness measurement of a moving system, as same in [5]. Experiment on SDP model shows the good performance of the proposed method in measuring collective. In future work, we will compute the collectiveness of various types of crowd systems, and provides useful information in crowd monitoring.